**Title**

Computer-aided diagnosis of lung nodule using gradient tree boosting and Bayesian optimization

**Authors and Affiliations**

Mizuho Nishio, M.D., Ph.D.[1,2*], Mitsuo Nishizawa, M.D., Ph.D.[3], Osamu Sugiyama, Ph.D.[2], Ryosuke Kojima, Ph.D.[4], Masahiro Yakami, M.D., Ph.D.[1,2], Tomohiro Kuroda, Ph.D.[5], Kaori Togashi, M.D., Ph.D.[1]

[1] Department of Diagnostic Imaging and Nuclear Medicine, Kyoto University Graduate School of Medicine, 54 Kawahara-cho, Shogoin, Sakyo-ku, Kyoto, Kyoto 606-8507, Japan

[2] Preemptive Medicine and Lifestyle Disease Research Center, Kyoto University Hospital, 53 Kawahara-cho, Shogoin, Sakyo-ku, Kyoto, Kyoto 606-8507, Japan

[3] Department of Radiology, Osaka Medical College, 2-7 Daigaku-machi, Takatsuki, Osaka 569-8686, Japan

[4] Department of Biomedical Data Intelligence, Kyoto University Graduate School of Medicine, 54 Kawahara-cho, Shogoin, Sakyo-ku, Kyoto, Kyoto 606-8507, Japan

[5] Division of Medical Information Technology and Administrative Plannnig, Kyoto University Hospital, 54 Kawahara-cho, Shogoin, Sakyo-ku, Kyoto, Kyoto 606-8507, Japan

**Abstract**

We aimed to evaluate computer-aided diagnosis (CADx) system for lung nodule classification focusing on (i) usefulness of gradient tree boosting (XGBoost) and (ii) effectiveness of parameter optimization using Bayesian optimization (Tree Parzen Estimator, TPE) and random search. 99 lung nodules (62 lung cancers and 37 benign lung nodules) were included from public databases of CT images. A variant of local binary pattern was used for calculating feature vectors. Support vector machine (SVM) or XGBoost was trained using the feature vectors and their labels. TPE or random search was used for parameter optimization of SVM and XGBoost. Leave-one-out cross-validation was used for optimizing and evaluating the performance of our CADx system. Performance was evaluated using area under the curve (AUC) of receiver operating characteristic analysis. AUC was calculated 10 times, and its average was obtained. The best averaged AUC of SVM and XGBoost were 0.850 and 0.896, respectively; both were obtained using TPE. XGBoost was generally superior to SVM. Optimal parameters for achieving high AUC were obtained with fewer numbers of trials when using TPE, compared with random search. In conclusion, XGBoost was better than SVM for classifying lung nodules. TPE was more efficient than random search for parameter optimization.

**Introduction**

Lung cancer is the leading cause of cancer deaths in the United States[1] because it is frequently diagnosed at an advanced stage and this prevents effective treatment. Results from the National Lung Screening Trial (NLST) show that compared with chest X-ray screening, lung cancer screening with low-dose CT significantly reduced lung cancer mortality among heavy smokers by detecting lung cancers at an early stage[2,3]. However, false positives in low-dose CT screening can be problematic and can result in unnecessary follow-up CT, positron emission tomography, or invasive procedures. In NLST, 96.4% of the positive results in the low-dose CT group were false positives[2,3].

Computer-aided diagnosis (CAD) has the potential of optimizing radiologists' workloads. CAD can assist radiologists in detecting lung nodules (CADe) and differentiating between benign and malignant nodules (CADx)[4-23]. CADx is useful for assisting radiologists in differentiating between benign and malignant lung nodules[6], and we expect it to be useful in reducing false positives in lung cancer screening with low-dose CT.

Gradient tree boosting is superior to off-the-shelf classifiers such as random forest or support vector machine (SVM)[24,25]. Because performance of CADx is affected by machine learning algorithms, gradient tree boosting may improve the performance of CADx. However, to the best of our knowledge, no other study has investigated the usefulness of gradient tree boosting in CADx of lung nodules. In our study, we used XGBoost as an implementation of gradient tree boosting[25] and applied it to CADx system of lung nodules.

It is necessary to optimize parameters of the machine learning algorithm to ensure good performance. Grid search has been frequently used for this purpose[26]. However, when the number of parameters is increased, grid search is not feasible because of its computational cost. As an alternative to grid search, random search and Bayesian optimization were used for parameter optimization[27,28]. Because XGBoost has many parameters, random search and Bayesian optimization are suitable for parameter optimization.

The purpose of the current study was to develop and evaluate the CADx system, focusing on (i) usefulness of XGBoost and (ii) effectiveness of parameter optimization using random search and Bayesian optimization. Herein, a hand-crafted imaging feature, a variant of the local binary pattern (LBP)[11,18,29-31], was used for calculating a feature vector that is fed into a machine learning algorithm.

**Methods**

This study used anonymized data from a public database. Regulations of Japan did not require institutional review board approval for use of a public database.

**CT images**

Our CADx system was tested using chest CT images obtained from The Cancer Imaging Archive (TCIA). TCIA is an open database of medical images, mainly consisting of CT, MRI, and nuclear medicine images that are stored as anonymized DICOM data. We used two sets of chest CT images from TCIA; one set from the LUNGx Challenge and one from the NSCLC Radiogenomics[20,21,32-35]. The LUNGx Challenge provided 60 test sets of chest CT images with 10 calibration sets. The 60 test sets included 73 lung nodules; a list of these nodules is available on the LUNGx Challenge website[34]. Among the 73 nodules from the LUNGx Challenge, 36 were lung cancers and 37 were benign. In NSCLC Radiogenomics, each of 26 sets of chest CT images included lung cancer. By combining data from LUNGx Challenge with NSCLC Radiogenomics, a total of 99 lung nodules (62 lung cancers and 37 benign nodules) were used for the development and evaluation of our CADx system.

**Image preprocessing**

First, CT images were loaded, and their voxel sizes were resampled into 1 × 1 × 1 mm. Next, the center was determined for each of the 99 nodules. Coordinates of the center of the lung nodules were provided via spreadsheet in the LUNGx Challenge and utilized here. Conversely, no such information was available for NSCLC Radiogenomics. Therefore, the center of the lung nodule was visually validated by two board-certified radiologists (M.N. and M.N.). A 64 × 64 × 64 3D bounding box was set for each nodule, and CT images inside the bounding box were cropped. The cropped 3D CT images were analyzed as the input to our CADx system. Areas of the CT images outside the bounding box were not assessed.

**Calculation of a feature vector**

The local binary pattern on three orthogonal planes (LBP-TOP) was used for calculating a feature vector[11,18,29,30,31]. Naïve implementation of 2D LBP was represented as follows:

$$LBP(x, R, P) = \sum_{i=0}^{P-1} 2^i \times s(d_i)$$

$$d_i = I\big(n(x, R, i)\big) - I(x),$$

where $x$ is the center pixel where LBP is calculated; $P$ is the number of samples; $n(x, R,$

*i*) is the *i*<sup>th</sup> neighbor pixel around the center pixel *x* and the distance between the center pixel *x* and the neighbor pixel is *R*; *I(u)* is the CT density of pixel *u* and *s(v)* is an indicator function, where *s(v)* is 1 if *v* ≥ 0 and 0 otherwise. We used a uniform pattern and rotation invariant type instead of naïve implementation as naïve implementation cannot handle large *P* values because they make feature vectors too long. Both uniform pattern and rotation invariant type can enhance the robustness of LBP as a feature vector. To utilize LBP in 3D CT images, LBP-TOP was used in this study. In LBP-TOP, 2D LBP was calculated on the XY, XZ and YZ planes and the texture information on other 3D planes was ignored. Then, the results of 2D LBP on the XY, XZ and YZ planes were converted into histograms, which were concatenated. 3D cropped CT images were evaluated with uniform pattern and rotation invariant type of LBP-TOP, and 1D feature vectors were calculated.

**Machine learning algorithm**

Our CADx system was built using SVM or XGBoost[24,25]. Implementations of SVM and XGBoost were freely available. SVM or XGBoost were trained using the feature vector obtained by LBP-TOP and its corresponding label. SVM is a widely used machine learning algorithm, and we used SVM with kernel trick (radial basis function) in this study. XGBoost builds an efficient classifier using gradient tree boosting. Gradient tree boosting is invariant to scaling of a feature vector, and it can determine higher-order interaction between a feature vector. Gradient tree boosting is trained in an additive manner. At each time step *t*, it grows another new tree to minimize the residual of the current model. Formally, the objective function of XGBoost can be described as follows:

$$L^t = \sum_{i=1}^{n} l\left(y_i, y_i^{t-1} + f_t(x_i)\right) + \Omega(f_t),$$

where $x_i$ and $y_i$ are the feature vector and its label at the $i^{th}$ instance, $y_i^{t-1}$ is the prediction of the $i^{th}$ instance at the $t - 1^{th}$ iteration, $f_t$ is the new tree that classifies the $i^{th}$ instance using $x_i$, $l$ denotes a loss function that measures the difference between the label and the prediction at the last step plus the new tree output, and $\Omega$ is the regularization term that penalizes the complexity of the new tree.

**Parameters**

The following parameter space was used for parameter optimization.

- For SVM, $C$ and $\gamma$ were used for controlling SVM with a radial basis function kernel. The range of $C$ and $\gamma$ were as follows: $C$, $1.0 \times 10^{-5}$–$1.0 \times 10^{5}$ and $\gamma$, $1.0 \times 10^{-5}$–$1.0 \times 10^{5}$.
- For XGBoost, parameters and their range were as follows: eta, 0.2–0.6; max_depth, 1–13; min_child_weight, 1–10; gamma, 0–1; learning_rate, $1.0 \times 10^{-4}$–$1.0 \times 10^{-1}$.
- LBP-TOP has two parameters ($R$ and $P$). The values of $R$ and $P$ were as follows: $R = 7, 8$ and $P = 40, 48$.

**Parameter optimization**

The parameter space was defined in the previous subsection. Here we denoted the parameters as $\boldsymbol{\theta}$. When using machine learning algorithm $A$ (SVM or XGBoost) and the parameter $\boldsymbol{\theta}$, we trained $A$ using training data and validated its performance using validation data. We used $L(A, \boldsymbol{\theta}, D_{train}, D_{valid})$ to denote the validation loss that $A$ achieved on validation data $D_{valid}$ when $A$ was trained on $\boldsymbol{\theta}$ and $D_{train}$. The parameter optimization problem under $K$-fold cross-validation was then to minimize the black box function:

$$f(\boldsymbol{\theta}) = \frac{1}{K}\sum_{i=1}^{K} L(A, \boldsymbol{\theta}, D_{train}^{i}, D_{valid}^{i}),$$

where $D_{train}^{i}$ and $D_{valid}^{i}$ were training data and validation data of the *i-th* fold of $K$-fold cross-validation, respectively. Bayesian optimization was used for optimizing this black box function $f(\boldsymbol{\theta})$ and for searching for the optimal parameter $\boldsymbol{\theta}$. Tree Parzen Estimator (TPE) was utilized for solving this problem[27]. Random search was used to compare the performance of TPE. Number of trials for TPE or random search was as follows: 10, 100, 200, and 1000.

**Statistical analysis**

Leave-one-out cross-validation was used for optimizing and evaluating the performance of our CADx system. Validation loss under leave-one-out cross-validation was used for parameter optimization. After parameter optimization, probabilistic outputs of our CADx system with optimal parameters were analyzed using accuracy and area under the curve (AUC) of receiver operating characteristic analysis. Classification results of our CADx system were output as probabilities of lung cancer to calculate AUC. For each number of trial, AUC and accuracy were calculated 10 times, and their averages were obtained. An outline of our CADx system is shown in Fig. 1.

**Results**

The averaged validation loss, AUC, and accuracy of our CADx system are shown in Tables 1 and 2 and Fig. 2–4. Supplementary Table S1 shows raw results of validation loss, AUC, and accuracy of our CADx system for each setting. Tables 1 and 2 show the averages of the raw results listed in Supplementary Table S1. Comparing the results depicted in Tables 1 and 2, XGBoost was generally superior to SVM. According to Table 1, the best averaged AUC of SVM was 0.850 when using TPE and number of trials = 1000. Table 2 shows that the best averaged AUC of XGBoost was 0.896 when using TPE and number of trials = 1000. According to Supplementary Table S1, the best AUC and accuracy of SVM was 0.855 and 0.834, respectively, and the best AUC and accuracy of XGBoost was 0.903 and 0.859, respectively. In XGBoost, the averaged AUC of TPE was better than that of random search when the number of trials was 100, 200, or 1000. In SVM, the averaged AUC of TPE was better than that of the random search when the number of trials was 100. However, when the number of trials was 10, the difference of the averaged AUC was minimal between random search and TPE in SVM and XGBoost. In addition, in SVM, the difference of averaged AUC was minimal between random search and TPE when the number of trials was 200 or 1000.

**Discussion**

In this study, we used two different sets of CT images for evaluating our CADx system; one set from the LUNGx Challenge and the other from the NSCLC Radiogenomics. The results of our CADx system show the following two main points; (i) XGBoost was better than SVM, and (ii) parameter optimization with TPE was better than that with random search. Using XGBoost and TPE, the best averaged AUC under leave-one-out cross-validation was 0.896 (the best AUC under leave-one-out cross-validation was 0.903).

Armato et al. performed an observer study using the 73 lung nodules of LUNGx Challenge, and the observer study included six radiologists. The results showed that AUC values of the six radiologists ranged from 0.70 to 0.85[21]. Although results of our CADx system were obtained using two sets of CT images (LUNGx Challenge and NSCLC Radiogenomics), we speculated that the diagnostic accuracy of our CADx system was comparable to that of the radiologists with respect to classifying lung nodules. Our CADx system might overfit the dataset of the current study. However, we speculated that the possibility of overfitting was not so high because this dataset consisted of two different sets of CT images, and the conditions and parameters of the images were variable (i.e. variability in use of contrast material and thickness of CT images).

The previous study shows that AUC value of CADx system was more than 0.8 by using the 73 lung nodules of LUNGx Challenge and SVM with a linear kernel[11]. Because of differences in CT images, quality of labels, and kernel type of SVM, it is difficult to precisely compare the diagnostic accuracy of the CADx system between the current study and the previous study. However, the diagnostic accuracy of our CADx system using SVM might be comparable to that of the previous study.

A few previous studies have utilized XGBoost for developing a clinical model. One study used XGBoost for classifying symptom severity based on text information in the form of psychiatrist notes[36]. Another study showed the usefulness of XGBoost for differentiation between subjects with epilepsy and healthy subjects using patients' cerebral activity assessed by functional MRI[37]. In conjunction with the results of these studies, we found that XGBoost was useful for developing an efficient and reliable clinical model. Although SVM was widely used as a machine learning algorithm in CADx, in our study, AUC of CADx using XGBoost was better than that using SVM. A prime reason for the superiority of XGBoost to SVM is invariant to scaling of a feature vector. As well-known kernels for SVM, such as radial basis function and linear kernels, are scale dependent, the output value of SVM is affected by scaling of a feature vector.

Previous studies have shown that Bayesian optimization was useful in several domains of clinical application[38-41]. The results of the current study are compatible with those of the previous studies. Figures 2–4 show that, in general, TPE is better than random search for optimizing parameters in SVM and XGBoost. However, when the number of trials was 10, the difference in performance between TPE and random search was minimal. This result suggests that the small number of trials (10) hindered parameter optimization of SVM and XGBoost. When the number of trials was 200 or 1000 in SVM, the difference in performance between random search and TPE was also minimal. Because parameter space of SVM was narrower than that of XGBoost in the current study, we surmised that both random search and TPE could almost fully optimize parameters and the difference in performance may be minimal.

There were several limitations to our study. First, the number of lung nodules was relatively small. Comparing the results of the current study and those of Armato et al.[21], we theorized that the diagnostic accuracy of our CADx system was comparable to that of radiologists for classifying lung nodules. However, this speculation might be optimistic as there is a possibility that our CADx system overfitted the dataset of the current study. Future studies should be conducted using a large number of lung nodules to prevent overfitting and evaluate the generalizability of our CADx system. Second, this study focused on the investigation of technical usefulness of XGBoost and Bayesian optimization from the viewpoint of CADx of lung nodules, and we ignored the clinical usefulness of our CADx system. Because the results of our study showed that the diagnostic ability of our CADx system may be comparable to that of radiologists, we expect that our CADx system will be useful for classifying lung nodules in a practical clinical setting. Third, the parameter space was relatively limited in this study. The parameters of our study were divided into two types: the parameter of the machine learning algorithm (i.e. *C* for SVM and eta for XGBoost) and the parameter of feature vectors (*R* and *P* of LBP). Because the results of parameter optimization were not stable when the parameter space of feature vectors was wide, we restricted the parameter space of feature vectors in our study. Last, we did not compare our CADx system with CADx using deep learning. Results of recent studies suggest that deep learning is superior to conventional machine learning. Therefore, our CADx system might be inferior to a CADx system with deep learning. We plan to develop a CADx system with deep learning and will use TPE for parameter optimization of deep learning in a future study.

In conclusion, XGBoost was better than SVM for classifying lung nodules. For optimizing parameters of both SVM and XGBoost, Bayesian optimization was more efficient than random search. Although our results were preliminary, the diagnostic accuracy of our CADx system may be comparable to that of radiologists for classifying lung nodules.

**Acknowledgements**

This study was supported by JSPS KAKENHI (Grant Number JP16K19883).

**Tables**

Table 1

Results of CADx when using SVM and parameter optimization

| Algorithm | Number of trial | Validation loss | AUC | Accuracy |
|---|---|---|---|---|
| Random | 10 | 0.528 | 0.792 | 0.734 |
| Random | 100 | 0.481 | 0.832 | 0.780 |
| Random | 200 | 0.460 | 0.848 | 0.794 |
| Random | 1000 | 0.451 | 0.849 | 0.789 |
| TPE | 10 | 0.515 | 0.797 | 0.724 |
| TPE | 100 | 0.461 | 0.847 | 0.802 |
| TPE | 200 | 0.458 | 0.846 | 0.792 |
| TPE | 1000 | 0.453 | 0.850 | 0.797 |

Abbreviation: computer-aided diagnosis, CADx; support vector machine, SVM; Tree Parzen Estimator, TPE; area under the curve, AUC.

Table 2

Results of CADx when using XGBoost and parameter optimization

| Algorithm | Number of trial | Validation loss | AUC | Accuracy |
| --- | --- | --- | --- | --- |
| Random | 10 | 0.488 | 0.838 | 0.756 |
| Random | 100 | 0.451 | 0.864 | 0.771 |
| Random | 200 | 0.440 | 0.868 | 0.784 |
| Random | 1000 | 0.422 | 0.878 | 0.806 |
| TPE | 10 | 0.494 | 0.838 | 0.762 |
| TPE | 100 | 0.427 | 0.876 | 0.811 |
| TPE | 200 | 0.419 | 0.881 | 0.804 |
| TPE | 1000 | 0.394 | 0.896 | 0.820 |

Abbreviation: computer-aided diagnosis, CADx; support vector machine, SVM; Tree Parzen Estimator, TPE; area under the curve, AUC.

**Figure and Figure legends**

Fig. 1. Outline of our CADx system

Abbreviations: CADx, computer-aided diagnosis; LBP-TOP, local binary pattern on three orthogonal planes; SVM, support vector machine

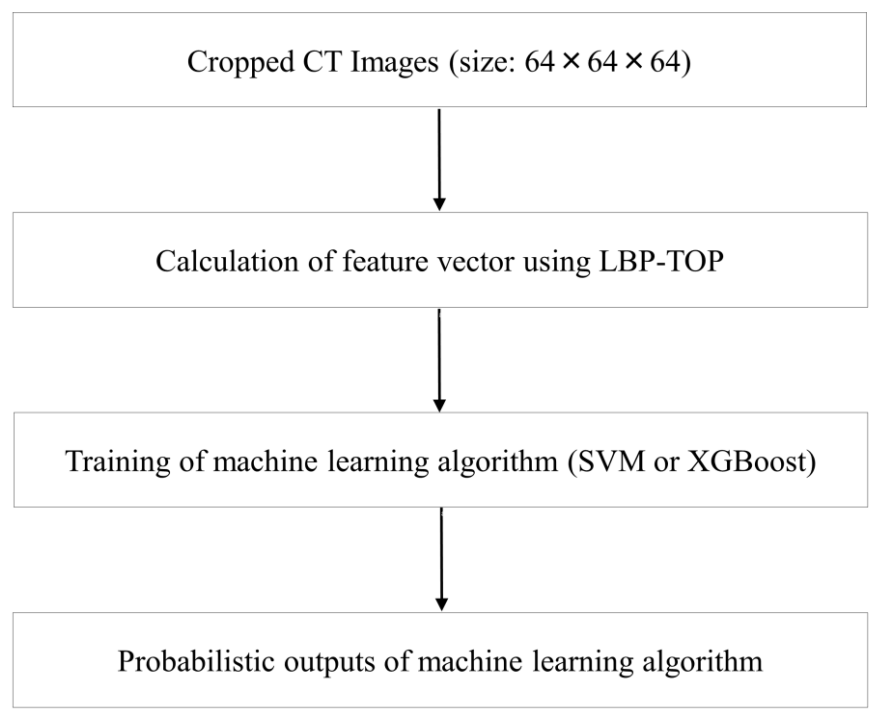

Fig. 2. Validation loss of CADx

Abbreviations: CADx, computer-aided diagnosis; SVM, support vector machine

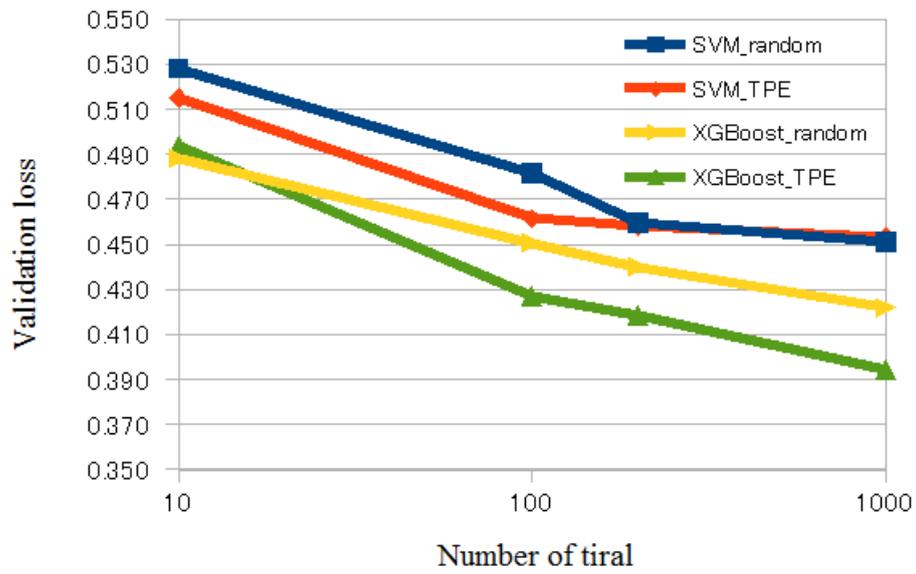

Fig. 3. AUC of CADx

Abbreviations: CADx, computer-aided diagnosis; SVM, support vector machine; AUC, area under the curve

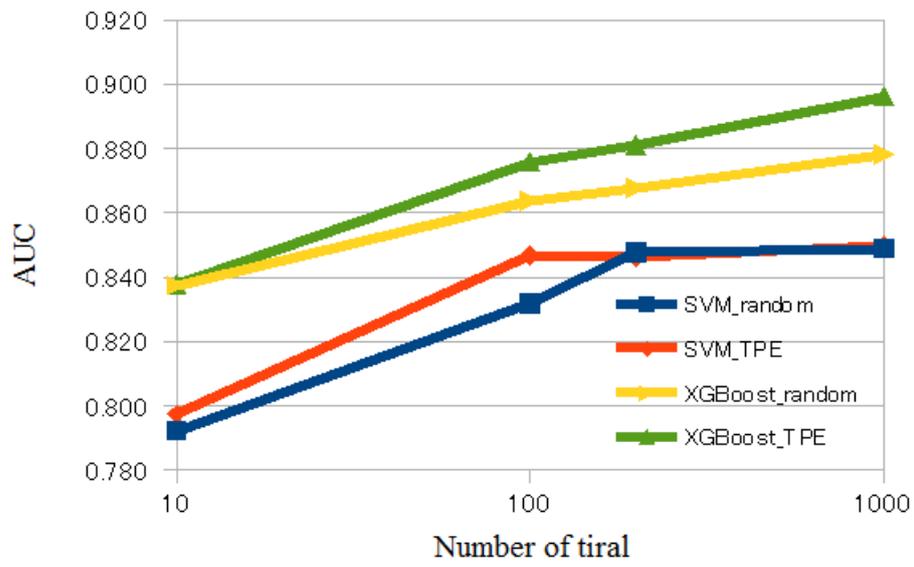

Fig. 4. Accuracy of CADx

Abbreviations: CADx, computer-aided diagnosis; SVM, support vector machine

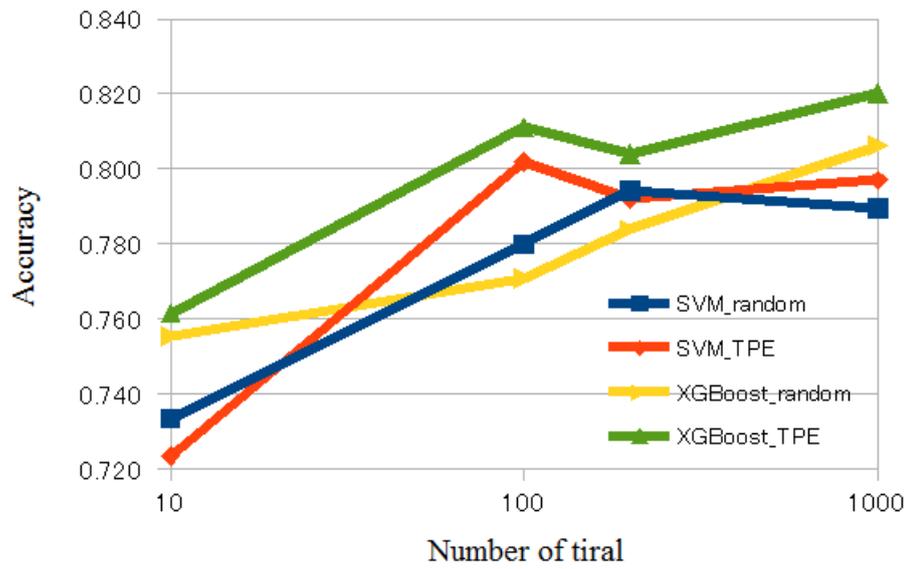

**Supplementary Information**

Supplementary Table S1. Raw results of parameter optimization.

Supplementary Table S1 shows raw results of validation loss, AUC, and accuracy of our CADx system for each setting.

**SVM**

| Algorithm | Number of trial | Validation loss | AUC | Accuracy |
|---|---|---|---|---|
| Random | 10 | | | |
| | | 0.472852459 | 0.838905841 | 0.816161616 |
| | | 0.467337894 | 0.847689625 | 0.786868687 |
| | | 0.494684872 | 0.828814298 | 0.796969697 |
| | | 0.52000721 | 0.785483871 | 0.718181818 |
| | | 0.528832919 | 0.782279861 | 0.67979798 |
| | | 0.564000536 | 0.759873583 | 0.739393939 |
| | | 0.579100753 | 0.749498692 | 0.644444444 |
| | | 0.590203773 | 0.734982563 | 0.651515152 |
| | | 0.561277597 | 0.771403662 | 0.722222222 |
| | | 0.501580928 | 0.823931997 | 0.77979798 |
| Random | 100 | | | |
| | | 0.453996306 | 0.84901918 | 0.782828283 |
| | | 0.486465281 | 0.824498692 | 0.758585859 |
| | | 0.463673017 | 0.85494769 | 0.782828283 |
| | | 0.50605269 | 0.810200523 | 0.766666667 |
| | | 0.466599725 | 0.841695728 | 0.818181818 |
| | | 0.491819683 | 0.827550131 | 0.786868687 |
| | | 0.478281981 | 0.83777245 | 0.772727273 |
| | | 0.500108052 | 0.816455972 | 0.758585859 |
| | | 0.499181226 | 0.812925022 | 0.75959596 |
| | | 0.468416559 | 0.842676548 | 0.813131313 |
| Random | 200 | | | |
| | | 0.449465093 | 0.850196164 | 0.802020202 |
| | | 0.462081879 | 0.846599826 | 0.816161616 |
| | | 0.459725217 | 0.846752398 | 0.818181818 |
| | | 0.458060719 | 0.845727986 | 0.780808081 |
| | | 0.466087248 | 0.85065388 | 0.792929293 |
| | | 0.459916646 | 0.842720139 | 0.782828283 |
| | | 0.453127683 | 0.846272886 | 0.787878788 |
| | | 0.4727752 | 0.843221447 | 0.778787879 |
| | | 0.466469114 | 0.852462947 | 0.770707071 |
| | | 0.448841497 | 0.853160418 | 0.811111111 |
| Random | 1000 | | | |
| | | 0.453286279 | 0.846534438 | 0.786868687 |
| | | 0.450903419 | 0.848997384 | 0.791919192 |
| | | 0.450341396 | 0.8494551 | 0.791919192 |
| | | 0.449242449 | 0.8505449 | 0.786868687 |
| | | 0.451700382 | 0.848256321 | 0.784848485 |
| | | 0.449170274 | 0.849585876 | 0.794949495 |
| | | 0.452971763 | 0.848953793 | 0.78989899 |
| | | 0.45200431 | 0.849585876 | 0.785858586 |
| | | 0.452164666 | 0.848125545 | 0.785858586 |
| | | 0.450675727 | 0.849476896 | 0.795959596 |
| TPE | 10 | | | |
| | | 0.517580418 | 0.788600697 | 0.686868687 |
| | | 0.529115386 | 0.785876199 | 0.694949495 |
| | | 0.515797145 | 0.809568439 | 0.766666667 |
| | | 0.550014703 | 0.754163034 | 0.670707071 |
| | | 0.516836166 | 0.784045336 | 0.675757576 |
| | | 0.529077427 | 0.775915432 | 0.719191919 |
| | | 0.454394539 | 0.849629468 | 0.790909091 |
| | | 0.532519961 | 0.795880558 | 0.737373737 |
| | | 0.49796965 | 0.830122058 | 0.762626263 |
| | | 0.503909225 | 0.79978204 | 0.731313131 |
| TPE | 100 | | | |
| | | 0.473905684 | 0.848910201 | 0.824242424 |
| | | 0.453385082 | 0.846730602 | 0.796969697 |
| | | 0.477545741 | 0.850741064 | 0.821212121 |
| | | 0.457103762 | 0.845836966 | 0.811111111 |
| | | 0.45151311 | 0.8505449 | 0.797979798 |
| | | 0.454227321 | 0.845989538 | 0.787878788 |
| | | 0.457181199 | 0.847210113 | 0.809090909 |
| | | 0.449047705 | 0.852244987 | 0.798989899 |
| | | 0.453109634 | 0.849171752 | 0.794949495 |
| | | 0.487489092 | 0.829555362 | 0.777777778 |
| TPE | 200 | | | |
| | | 0.449103032 | 0.851198779 | 0.78989899 |
| | | 0.450863972 | 0.84989102 | 0.792929293 |
| | | 0.449983732 | 0.851438535 | 0.787878788 |
| | | 0.452206302 | 0.848038361 | 0.78989899 |
| | | 0.451227635 | 0.84923714 | 0.793939394 |
| | | 0.486096436 | 0.834459459 | 0.8 |
| | | 0.450834341 | 0.851002616 | 0.807070707 |
| | | 0.45350115 | 0.846512642 | 0.801010101 |
| | | 0.452603367 | 0.846948561 | 0.785858586 |
| | | 0.482127102 | 0.831560593 | 0.771717172 |
| TPE | 1000 | | | |
| | | 0.448719475 | 0.852484743 | 0.814141414 |
| | | 0.451453543 | 0.849520488 | 0.793939394 |
| | | 0.451718977 | 0.848866609 | 0.793939394 |
| | | 0.478046461 | 0.849040976 | 0.834343434 |
| | | 0.451485383 | 0.849716652 | 0.782828283 |
| | | 0.448444877 | 0.851002616 | 0.788888889 |
| | | 0.450218994 | 0.848605057 | 0.783838384 |
| | | 0.449448807 | 0.848234525 | 0.784848485 |
| | | 0.451326652 | 0.850479512 | 0.787878788 |
| | | 0.449786465 | 0.851830863 | 0.806060606 |

**XGboost**

| Algorithm | Number of trial | Validation loss | AUC | Accuracy |
|---|---|---|---|---|
| Random | 10 | | | |
| | | 0.445145341 | 0.865736704 | 0.797979798 |
| | | 0.482081523 | 0.838709677 | 0.777777778 |
| | | 0.490143274 | 0.841325196 | 0.757575758 |
| | | 0.454091127 | 0.860505667 | 0.808080808 |
| | | 0.527766411 | 0.806887533 | 0.737373737 |
| | | 0.467754416 | 0.848735833 | 0.757575758 |
| | | 0.521778064 | 0.810810811 | 0.686868687 |
| | | 0.471237447 | 0.846556234 | 0.797979798 |
| | | 0.539348503 | 0.817349608 | 0.717171717 |
| | | 0.483087281 | 0.838709677 | 0.717171717 |
| Random | 100 | | | |
| | | 0.446691798 | 0.869659983 | 0.767676768 |
| | | 0.436859639 | 0.869224063 | 0.787878788 |
| | | 0.473479152 | 0.850043592 | 0.717171717 |
| | | 0.458098385 | 0.850043592 | 0.777777778 |
| | | 0.446666752 | 0.857454228 | 0.757575758 |
| | | 0.44378296 | 0.868352223 | 0.777777778 |
| | | 0.458026174 | 0.857454228 | 0.755575758 |
| | | 0.445379934 | 0.872711421 | 0.757575758 |
| | | 0.452673391 | 0.869659983 | 0.797979798 |
| | | 0.444972962 | 0.870531822 | 0.808080808 |
| Random | 200 | | | |
| | | 0.443760611 | 0.863121186 | 0.767676768 |
| | | 0.444146966 | 0.868788143 | 0.777777778 |
| | | 0.424888947 | 0.87532694 | 0.777777778 |
| | | 0.445441904 | 0.857018309 | 0.787878788 |
| | | 0.446177996 | 0.861813426 | 0.767676768 |
| | | 0.437196368 | 0.869224063 | 0.797979798 |
| | | 0.461010096 | 0.858326068 | 0.797979798 |
| | | 0.430482423 | 0.877506539 | 0.787878788 |
| | | 0.429184647 | 0.87401918 | 0.777777778 |
| | | 0.437877671 | 0.872275501 | 0.797979798 |
| Random | 1000 | | | |
| | | 0.434565969 | 0.862685266 | 0.808080808 |
| | | 0.387608598 | 0.894507411 | 0.838383838 |
| | | 0.437005104 | 0.870531822 | 0.757575758 |
| | | 0.431226879 | 0.878814298 | 0.828282828 |
| | | 0.418224606 | 0.883609416 | 0.808080808 |
| | | 0.426340077 | 0.873583261 | 0.828282828 |
| | | 0.431543871 | 0.878378378 | 0.797979798 |
| | | 0.412123485 | 0.884481255 | 0.808080808 |
| | | 0.419350243 | 0.877506539 | 0.808080808 |
| | | 0.419302571 | 0.876198779 | 0.777777778 |

| TPE | 10 | | | |
|---|---|---|---|---|
| | | 0.500940918 | 0.83391456 | 0.777777778 |
| | | 0.437500892 | 0.865736704 | 0.787878788 |
| | | 0.486661284 | 0.832170881 | 0.747474747 |
| | | 0.482096219 | 0.841761116 | 0.767676768 |
| | | 0.4862092 | 0.836094159 | 0.755757575 |
| | | 0.501471978 | 0.843504795 | 0.755757575 |
| | | 0.501419561 | 0.827375763 | 0.737373737 |
| | | 0.566848912 | 0.796425458 | 0.747474747 |
| | | 0.491636458 | 0.841325196 | 0.767676768 |
| | | 0.480854742 | 0.858326068 | 0.767676768 |
| TPE | 100 | | | |
| | | 0.399637731 | 0.897994769 | 0.858585859 |
| | | 0.441848314 | 0.872275501 | 0.787878788 |
| | | 0.445585612 | 0.867044464 | 0.797979798 |
| | | 0.435037646 | 0.868352223 | 0.797979798 |
| | | 0.429655877 | 0.877942459 | 0.808080808 |
| | | 0.397312074 | 0.891455972 | 0.828282828 |
| | | 0.44651141 | 0.859633827 | 0.808080808 |
| | | 0.417638445 | 0.880122058 | 0.828282828 |
| | | 0.435326973 | 0.870967742 | 0.767676768 |
| | | 0.420929913 | 0.871403662 | 0.828282828 |
| TPE | 200 | | | |
| | | 0.397735901 | 0.903225806 | 0.818181818 |
| | | 0.436144729 | 0.868788143 | 0.767676768 |
| | | 0.405300595 | 0.886660854 | 0.848484848 |
| | | 0.423472902 | 0.873583261 | 0.797979798 |
| | | 0.428297094 | 0.880557977 | 0.787878788 |
| | | 0.41926367 | 0.884045336 | 0.808080808 |
| | | 0.416707359 | 0.879250218 | 0.797979798 |
| | | 0.425973878 | 0.872275501 | 0.808080808 |
| | | 0.428851232 | 0.87401918 | 0.777777778 |
| | | 0.404465247 | 0.889276373 | 0.828282828 |
| TPE | 1000 | | | |
| | | 0.402657802 | 0.89581517 | 0.777777778 |
| | | 0.378864325 | 0.900174368 | 0.858585859 |
| | | 0.379654844 | 0.901918047 | 0.858585859 |
| | | 0.404564335 | 0.894071491 | 0.777777778 |
| | | 0.380837913 | 0.900610288 | 0.858585859 |
| | | 0.420429095 | 0.881865737 | 0.808080808 |
| | | 0.383862273 | 0.897558849 | 0.848484848 |
| | | 0.403646672 | 0.894507411 | 0.777777778 |
| | | 0.384985044 | 0.900610288 | 0.858585859 |
| | | 0.404562696 | 0.894071491 | 0.777777778 |